\documentclass{article}
\usepackage{spconf,amsmath,graphicx}
\usepackage{arabtex}
\usepackage{utf8}

\usepackage{booktabs}
\usepackage{amssymb}
\usepackage{amsfonts}
\usepackage{multicol}
\usepackage{multirow}
\usepackage[colorinlistoftodos]{todonotes}
\usepackage{hyperref}

\usepackage{arabtex}
\usepackage{utf8}

\usepackage{times}
\usepackage{url}
\usepackage{latexsym}
\usepackage{graphicx}
\usepackage{float}
\restylefloat{table}
\usepackage{enumitem}
\usepackage{color,soul}
\usepackage{verbatim}

\setcode{utf8}

\title{WERd: Using Social Text Spelling Variants\\ for Evaluating Dialectal Speech Recognition}
\makeatletter
\def\name#1{\gdef\@name{#1\\}}
\makeatother \name{Ahmed Ali$^{1,2}$, Preslav Nakov$^1$, Peter Bell$^2$, Steve Renals$^2$}

\address{$^1$Qatar Computing Research Institute, HBKU, Doha, Qatar \\
$^2$Centre for Speech Technology Research,
University of Edinburgh, UK \\
  {\footnotesize \{amali, pnakov\}@qf.org.qa, \{peter.bell, s.renals\}@ed.ac.uk}
}
\begin{document}
\maketitle
\begin{abstract}
We study the problem of evaluating automatic speech recognition (ASR) systems that target dialectal speech input. 
A major challenge in this case is that the orthography of dialects is typically not standardized.
From an ASR evaluation perspective, 
this means that there is no clear gold standard for the expected output, and several possible outputs could be considered correct according to different human annotators, which makes standard word error rate (WER) inadequate as an evaluation metric. 
Such a situation is typical for machine translation (MT), and thus we borrow ideas from an MT evaluation metric, namely TERp, an extension of translation error rate which is closely-related to WER. In particular, in the process of comparing a hypothesis to a reference, we make use of spelling variants for words and phrases, which we mine from Twitter in an unsupervised fashion. Our experiments with evaluating ASR output for Egyptian Arabic, and further manual analysis, show that the resulting WERd (i.e.,  \emph{WER for dialects}) metric, a variant of TERp, is more adequate than WER for evaluating dialectal ASR.
\end{abstract}


\section{Introduction}

Automatic Speech Recognition (ASR) has shown fast progress recently, thanks to advancements in deep learning.
As a result, 
the best systems for English have achieved a single-digit word error rate (WER) for some conversational tasks~\cite{saon2017english}.
However, this is different for dialectal ASR, for which the WER can easily go over 40\%\cite{aliMGB3}.


In a standardized language such as English, we know that \emph{enough} is a correct spelling, while \emph{enuf} is not. However, we cannot be sure about the correct spellings of dialectal words; at best, we would know what a preferred or a dominant spelling is. 
This is because dialects typically do not have an official status and thus their spelling is not regulated, which opens widely the door to orthographic variation.\footnote{Note that here we target primarily intra-dialectal variation. Yet, there is also inter-dialect variation, e.g., between the different dialects of Arabic.} 

\begin{table}[!htbp]
\small
\centering
\begin{tabular}{lcc}
\bf English Gloss & \bf Spelling Variants & \bf Buckwalter \\
\hline
He was not & \<ماكانش> & mAkAn\$\\
& \<ماكنش> & mAkn\$\\
& \<ما كانش> & mA kAn\$\\
& \<مكنش> & mkn\$\\
\hline
I told him & \<قولتله> & qwltlh\\
& \<قولت له> & qwlt lh\\
& \<قلتله> & qltlh\\
& \<قلت له> & qlt lh\\
\hline
By the morning & \<على الصبح> & ElY AlSbH\\
& \<علي الصبح> & Ely AlSbH\\
& \<ع الصبح> & E AlSbH\\
& \<عالصبح> & EAlSbH\\
& \<عصّبح> & ESbH\\
\hline
\end{tabular}
\caption{\label{tab:samples}Egyptian phrases with multiple spelling variants: shown in Arabic script and in Buckwalter transliteration.}
\end{table}

\noindent Table~\ref{tab:samples} shows some examples of spelling variation in Dialectal Arabic
(DA). We can see that clitics (pronouns and negations) can be written concatenated or separated from the verb, the definite article can undergo different spelling variations due to coarticulation with the following word, long vowels can become short, and thus be dropped as they are typically not written in Arabic, etc.
While some variations can happen in standardized languages such as English, e.g., \emph{healthcare} vs. \emph{health care}, or \emph{organize} vs. \emph{organise}, this is much less common, and in ASR it is easily handled with simple rules, e.g., the Global Mapping file\footnote{The global mapping file 
can help for handcrafted variants like \emph{color/colour} and \emph{ten/10} in English. However, it is not applicable to dialectal Arabic, where multiple spelling variants are acceptable; we use 11M pairs.}
in \textsc{sclite} \cite{rosenfeld1997cmu,fiscus1997post}.

The above examples partially explain the high WER for dialects. While they suffer from the lack of training resources, the main problem is their informal status, which means that their spelling is rarely regulated. This makes training an ASR system for dialects much harder as there is no single gold standard towards which to optimize at training time. 

\noindent More importantly, it is hard to evaluate such a system and to measure progress 
as multiple possible 
text outputs for the same speech signal could be considered correct by different people. 
Thus, there is need for an evaluation measure that would allow for common spelling variations. 
In this work, we propose to mine such variations from dialectal Arabic tweets and to incorporate them as spelling variants as part of a more adequate ASR evaluation measure for dialects.

Previously, the problem was addressed using the multi-reference word error rate (MR-WER) \cite{ali2015multi}, which is similar to the multi-reference BLEU score \cite{Papineni:2002:BLEU} used to evaluate Machine Translation (MT). However, obtaining multiple references is expensive. Moreover, it could take many human annotators to get good coverage of the possible orthographic variants of the transcription of a speech recording. Thus,  we propose to use a single reference, but to perform matching using spelling variants that could capture some of the variation. 

This was applied to MT, e.g., 
for parameter optimization \cite{madnani-EtAl:2007:WMT}, where additional synthetic references are generated for tuning purposes,
or for phrase-based SMT, where paraphrasing is applied to the source side of the phrase table \cite{Callison-Burch:al:2006:mt}, of the training bi-text \cite{nakov:2008:WMT}, or both \cite{Nakov:2008:ISM,Nakov:2011:TMC,Wang:2012:SLA,Wang:2016:SLA}.
Paraphrasing has been also used for evaluating text summarization \cite{Zhou:al:2006:mt}.

More relevant to the present work, in MT evaluation,
paraphrasing was applied to the output of an MT system \cite{Kauchak:Barzilay:2006:par}.
It was also incorporated in measures such as TERp \cite{Snover:2009:TERp}, which is a translation edit rate metric with paraphrases. Indeed, here we borrow ideas from TERp for dialectal ASR, with a paraphrase table (in our case, a spelling variants table), which we mine automatically from a huge collection of tweets in an unsupervised fashion. Our experiments and our manual analysis show that this is a very promising idea.

Our contributions are as follows:
(\emph{i}) We propose a method for automatically collecting spelling and tokenization variations for dialectical Arabic (and, presumably, other languages and language variants) from Twitter data;
(\emph{ii}) We further incorporate these spelling variants in an evaluation metric, WERd, which is variation of TERp, and we demonstrate its utility for dialectal Arabic ASR.
We release the code for that metric, as well as the spelling variants we mined and used in the metric:\footnote{\url{https://github.com/qcri/werd}} eleven million pairs, which we extracted from a seven-billion words corpus of dialectal Arabic tweets.

\section{Method}
\label{method}

We propose a method for evaluating dialectal ASR,
which consists of two steps: (\emph{i})~collecting a large number of spelling variants, which we mine from social media in an unsupervised manner, and (\emph{ii})~using these spelling variants, with associated probabilities, into an MT-inspired evaluation measure (together with standard unit-cost word insertions, deletions, and substitutions).

\begin{figure}[h!]
\centering
\includegraphics [scale=1.5]{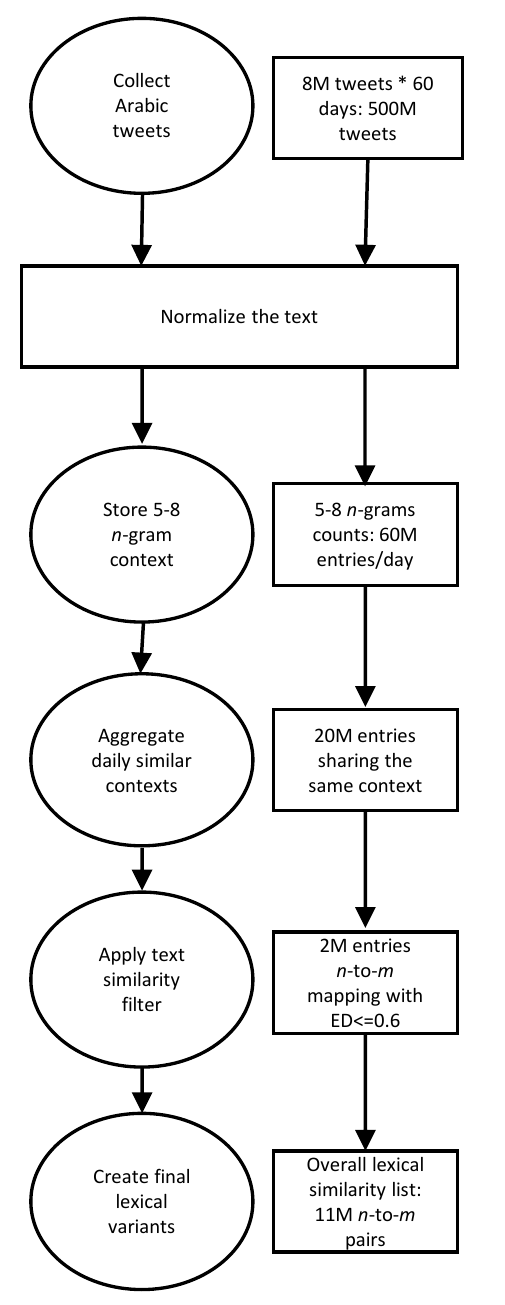}
\caption{\label{Fig:flowchart} Diagram of our pipeline for extracting dialectal spelling variants from Twitter.}
\end{figure}

\subsection{Mining Spelling Variants from Social Media}

We use social media to mine dialectal spelling variants from a collection of half a billion dialectal Arabic tweets.
Our approach is language-independent, scalable,
and unsupervised, as it assumes no prior knowledge about the language, its dialects, or the data. 

We build a list of pairs of spelling variants with probabilities using the following steps (as shown in Figure~\ref{Fig:flowchart}):

First, we collect Arabic tweets.
Then, we normalize hashtags, URLs, emoticons. We further drop Arabic diacritics and 
elongation, and we reduce letter repetitions to maximum three. Our pipeline is an extension to the previous work done in Arabic language processing for microblogs \cite{darwish2012language}.

Next, we extract all $n$-grams of lengths 5--8.
In each $n$-gram, we consider the first two and the last two words as a context, and the 1--4 words in the middle as a \emph{target} for this context.
For example, for a 5-gram we will have $<L_1, L_2, t_1, R_1, R_2>$, 
while for an 8-gram we will have $<L_1, L_2, t_1, t_2, t_3, t_4, R_1, R_2>$, where $L_i$ and $R_i$ represent the left and the right context words, and $t_j$ are the target words in the middle ($1 \leq i \leq 2$, $1 \leq j \leq 4$).

Next, we generate pairs of potential spelling variants for targets that share the same contexts.
This is subject to the constraint that the normalized Levenshtein distance between the targets is 
less than $t$, measured in characters.
We tried values between 0.1 and 0.6 for $t$, and we manually inspected the resulting pairs of spelling variants. Ultimately, we set $t=0.6$. With normalization in mind, we further impose a constraint that in each pair of spelling variants, one of the targets is extracted in the same contexts at least $N$ times more frequently than the other one (we set $N$ to 3).
Finally, with each pair of spelling variants, we associate a score: the average of the two Levenshtein distances. 
The resulting scored pairs of spelling variants form a spelling variant table for WERd. 

Here are two examples from this final table of \textit{n}-to-\textit{m} spelling variant pairs with corresponding frequencies and normalized edit distance (shown in Buckwalter): \newline 
\colorbox{yellow}{mAfy} \colorbox{green}{mAAfy}  \colorbox{pink}{752} \colorbox{red}{75} \colorbox{gray}{0.25}\newline
\colorbox{yellow}{lwny w DAEt} \colorbox{green}{lwny wDAEt}  \colorbox{pink}{32} \colorbox{red}{8} \colorbox{gray}{0.1}\newline

The first column (yellow) contains the frequent form, which is the target \textbf{mAfy}. The second column (green) contains the source \textbf{mAAfy}, which is a less frequent term. The next column is the frequency of the target, e.g., the word \textbf{mAfy} occurred 752 times. The following column is the frequency of the source in the same context, e.g., \textbf{mAAfy} occurred 75 times. Finally comes the normalized edit distance. 

Related approaches for paraphrase extraction have used random walks \cite{hassan2013social}, pairwise similarity \cite{han2012automatically}, and continuous representations \cite{sridhar2015unsupervised,sproat2016rnn}. Unlike that work, we mine pairs of spelling variants for ASR evaluation, not for modeling; we further allow many-to-many mappings, and we do not target canonical gold normalization. 


\subsection{Using the Spelling Variants for Evaluation: WERd}

We borrow ideas from an evaluation measure for MT evaluation, namely \emph{Translation Edit Rate Plus} or \emph{TERp} \cite{snover2006study}.
TERp allows block alignment of words, called \emph{shifts} within the hypothesis as a low cost edit, a cost of 1, the same as the cost for inserting, deleting or substituting a word. 
TERp uses a greedy search and shift constraints to both reduce the computational complexity and to model the quality of translation better. The metric further supports tuned weights for the edit operations, a paraphrase table, synonym/hypernym-based matching using WordNet, etc. 

The main motivation for using paraphrases in TERp for MT evaluation is to capture some lexical variation, e.g., \emph{(controversy over, polemic about), (by using power, by force), (brief, short), (response, reaction)}.
In contrast, we focus on capturing spelling variation in a dialect as shown in Table~\ref{tab:samples}.

In this work, we only use the paraphrasing capability of TERp. We restrict the matching to monotonic, i.e., no reorderings and no shifts.
The only additional operation that we allow, compared to WER, is mapping between the hypothesis and the reference using a pair of spelling variants from our spelling variants table, which can span up to four words on either side of the pair of spelling variants as we have explained above. This monotonic version of TERp, with no reordering but with spelling variant matching capabilities gives rise to our metric for dialectal ASR evaluation, which we will call WERd (or \emph{WER for dialects}).

\section{Experiments and Evaluation}

\subsection{Dialectal Data}

\textbf{Speech Data.} We collected two hours of Egyptian Arabic Broadcast news \cite{wray2015crowdsource} 
speech data, which we split into 1,217 segments, each 3-10 seconds long. 
It can be argued that Egyptian Arabic, which is one of the Arabic dialects, is a language with no established orthographic rules. This makes it difficult to develop standard transcription guidelines covering orthography. Therefore, we decided to have multiple transcriptions, but to let transcribers write the transcripts as they deemed correct, while trying to be as verbatim as possible. All the transcribers are native speakers of the chosen dialect with no linguistic background.\footnote{The transcribers were asked to follow these transcription guidelines: \url{http://alt.qcri.org/resources/MGB-3/Arabic_Transcription\%20_Guidelines_20170330.pdf}}
Table~\ref{tab:interanno} shows the overlap between the annotators, at the segment level, for their original transcription and after applying surface normalization for \emph{alef}, \emph{yah} and \emph{hah}, which is standard for Arabic. 
In Table~\ref{tab:interanno}, the first number is for the original text, and the second number is for the normalized text. 
We can see that even after normalization,\footnote{Below, we will report results after normalization only.} there are about 15\% differences between most of the annotators. 


%

\noindent \textbf{Social Media Data.} We further collected dialectal Arabic tweets in order to extract spelling variants. In particular, we issued queries using \texttt{lang:ar} against the Twitter API\footnote{\url{http://dev.twitter.com/}}. Note that we did not try to control the location where the tweets originated from, but only the language they were written in. We collected two months of tweets (from December 2015 and January 2016), with about eight million tweets per day on average, which yielded a total of half a billion tweets containing over seven billion word tokens.
\subsection{ASR System}




For our experiments, we used the speech-to-text transcription system built as part of QCRI's submission to the 2016 Arabic Multi-Dialect Broadcast Media Recognition (MGB-2) Challenge\cite{khurana2016qcri}.
Here are some key features of that system:
%
%

\textbf{Data:} The training data consisted of 1,200 hours of transcribed broadcast speech data collected from the Aljazeera news channel. In addition, we had ten hours of development data \cite{ali2016mgb}. We used data augmentation techniques such as speed and volume perturbation, which increased the size of the training data to three times the original size 
\cite{ko2015audio}.

\textbf{Speech lexicon:} We used a grapheme-based lexicon \cite{killer2003grapheme} with 900k entries, which we constructed using the words that occurred more than twice in the training transcripts.

\textbf{Speech features:} To train the acoustic models, we used 40-dimensional high-resolution Mel Frequency Cepstral Coefficients 
(MFCC\_hires), extracted for each speech frame, which we concatenated with 100-dimensional i-Vectors per speaker in order to facilitate speaker adaptation \cite{saon2013speaker}.

\textbf{Acoustic models:} We experimented with three acoustic models: Time-Delayed Neural Networks (TDNNs) \cite{peddinti2015time}, Long Short-Term Memory Recurrent Neural Networks (LSTMs), and Bi-directional LSTMs (BiLSTMs) \cite{sak2014long}. The latter acoustic model outperforms the other two models in terms of WER. We trained all models using Lattice-Free Maximum Mutual Information (LF-MMI) \cite{povey2016purely} using the Kaldi toolkit \cite{povey2011kaldi}.

\textbf{Language model:} We trained two $n$-gram language models (LMs). The first one, a tri-gram LM (KN3) used the spoken utterance transcripts for the 1,200 hours. We used this LM in order to generate decoding lattices. We then rescored these lattices using a four-gram LM (KN4), which we trained on the in-domain data and on some extra text. We used interpolated Kneser-Ney smoothing for both LMs, which we built using the SRILM toolkit \cite{stolcke2002srilm}. 
We further trained a Recurrent Neural Network Language Model with MaxEnt Connections (RNNME) using the RNNLM toolkit \cite{mikolov2011rnnlm}

\textbf{Overall ASR system:} We combined the three aforementioned acoustic models, and for the second pass we additionally used the four-gram and the RNNLM for re-scoring the decoded speech lattices. The overall performance was 14.7\% WER on the MGB-2 tasks, and this was the best result achieved at the challenge.

\begin{table}[t]
\begin{center}
\begin{tabular}{c|cccc}
 \multicolumn{1}{c}{ } & ref2 & ref3 & ref4 & ref5  \\
 \hline
 ref1 & 77/86 & 80/84 & 78/86 & 80/87 \\
 ref2 & --- & 74/83 & 71/85 & 72/85 \\
 ref3  & --- & --- & 77/84 & 78/84 \\
 ref4 & --- & --- & --- & 91/93 \\
 \hline
\end{tabular}
 \caption{\label{tab:interanno}Pairwise overlap of the five human references before/after normalization (in \%).}
\end{center}
\end{table}

\subsection{Experimental Results}

We first evaluated the ASR system on our two-hour dialectal Arabic test dataset using WER with respect to each of the five references. The results are shown in Table~\ref{tab:WER}. We can see that the WER is much higher on our dialectal Arabic dataset, ranging in 40--50\%.

We further calculated MR-WER for our ASR system using all five references, achieving a score of 25.3\%. This number is much lower than when evaluating with respect to any individual reference, which is to be expected, as we allow more matching options.



%

\begin{table}[tb]
\begin{center}
\begin{tabular}{ccccc}
 \multicolumn{1}{c}{ } & \multicolumn{1}{c}{\bf WER} & \multicolumn{1}{c}{\bf TER} & \multicolumn{1}{c}{\bf WERd} & \multicolumn{1}{c}{\bf MR-WER} \\
 \hline
 ref1 & 46.2 & 37.4 & 34.3 & ---\\ 
 ref2 & 42.9 & 38.7 & 35.7 & ---\\ 
 ref3 & 48.9 & 41.9 & 38.3 & --- \\ 
 ref4 & 46.2 & 39.0 & 35.6 & --- \\ 
 ref5 & 46.0 & 38.3 & 34.9 & --- \\ 
 \hline
 ALL refs & --- & --- & --- & 25.3\\
  \end{tabular}
 \caption{\label{tab:WER}WER vs. TER vs. WERd vs. MR-WER, after normalization (in \%).}
\end{center}
\end{table}

Table~\ref{tab:WER} reports TER and WERd scores calculated with respect to each of the five references and shows for both metrics a strong correlation with WER. We can also see that the scores for WERd are halfway between WER and MR-WER (e.g., for ref1, it is 34.3 vs. 46.2 and 25.3, respectively), but without the need for additional human references. 

\begin{table*}[tbh]
\begin{center}
\begin{tabular}{ cccccccc}
  &  \bf No Variants & \bf ED $\leq$ 0.1 & \bf ED $\leq$ 0.2 & \bf ED $\leq$ 0.3 & \bf ED $\leq$ 0.4 & \bf ED $\leq$ 0.5 & \bf ED $\leq$ 0.6 \\
 \hline
 ref1 & 37.4 & 37.1 & 36.6 & 35.5 & 34.8 & 34.6 & 34.3 \\
ref2 & 38.7 & 38.4 & 37.9 & 36.9 & 36.3 & 36.1 & 35.7 \\
ref3 & 41.9 & 41.5 & 40.9 & 39.7 & 38.9 & 38.7 & 38.3 \\
ref4 & 39.0 & 38.6 & 38.1 & 36.9 & 36.2 & 36.0 & 35.6 \\
ref5 & 38.3 & 37.9 & 37.3 & 36.2 & 35.6 & 35.3 & 34.9 \\
\hline
\end{tabular}
\end{center}
\caption{\label{tab:TERP}WERd using pairs of spelling variants extracted using different maximum edit distances (ED).}
\end{table*}

\begin{figure*}[tbh] \label{Fig:Eval_WERP}
\centering
\includegraphics [scale=0.6]{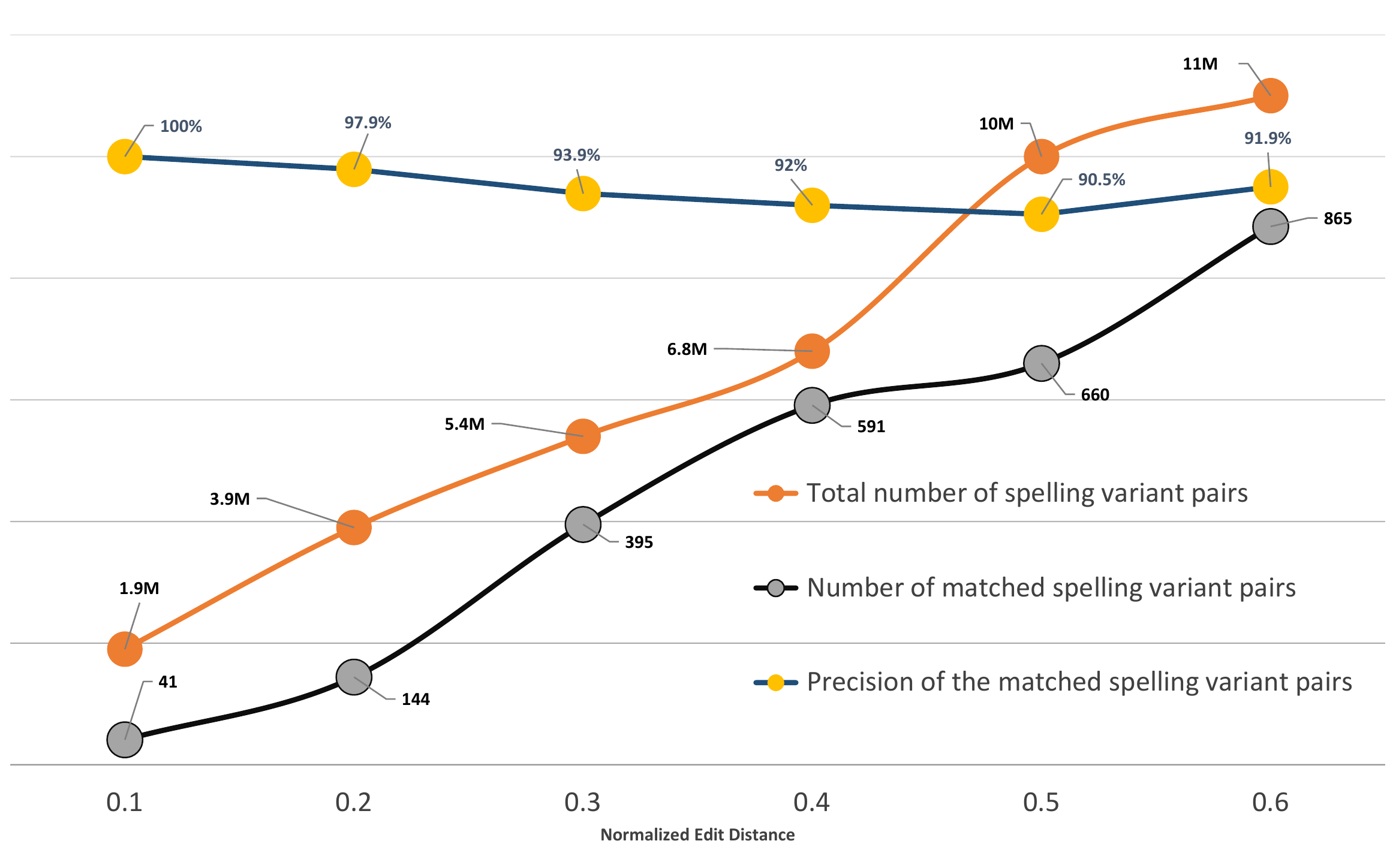}
\caption[caption]{Analysis of the total number of spelling variants, the number of matched variants, and the precision for different thresholds on the edit distance (with respect to ref1). Note that the $y$ axis shows different units for each of the three curves.}
\end{figure*}

There are two reasons for MR-WER to be considerably lower compared to the other metrics. First, the way foreign words and code-switching are handled by different annotators, 
e.g., words like \emph{BBC} can be written in either Arabic (\<بي بي سي> \emph{by by sy}) or Latin characters. Some annotators would use Arabic while other would prefer English, which would allow matching either of them when evaluating the ASR output with multiple references. Second, in dialectal Arabic, there are many filler words such as \<يعني> \emph{yEny},  \<اصل> \emph{ASl} and \<زي> \emph{zy}, which some annotators would skip and some would keep.

\section{Discussion}

\noindent \textbf{WERd for different thresholds.} Table~\ref{tab:TERP} shows the performance of WERd when using pairs of spelling variants with different maximum edit distances: 0.1--0.6. 
As the threshold increases, WERd decreases,
e.g., for ref1, it goes from 37.4 to 34.3, or 8\% relative reduction.
The difference is due to the number of matched spelling variant pairs, e.g., 865 for ref1.

\smallskip
\noindent
\textbf{Analysis of the pairs of spelling variant matches.} Next, we study the relationship between the threshold on the maximum edit distance vs. the spelling variant table size,  the number of spelling variants matches, and 
the accuracy of these matches.
This is shown in Figure~\ref{Fig:Eval_WERP}, where we focus the best reference, ref1 (according to native speakers of Egyptian Arabic who have a linguistic background).
We can see that the threshold has a major impact on the spelling variant table size: going from 0.1 to 0.6 
yields a six times larger table. It also yields a 21 times larger number of spelling variant matches on the test dataset: from 41 to 865. 

\noindent Of course, this comes at a cost: while all 41 spelling variant matches at threshold of 0.1 are correct, there are 8\% errors among the 865 
matches at threshold of 0.6.
We believe this is a relatively small price to pay, given the advantage of being able to identify 791 additional correct matches,
which we capture without the need for having multiple references.\footnote{We were unable to measure the recall as it requires manual evaluation of all the possible candidates for spelling variants in the references.}

\smallskip
\noindent
\textbf{Pearson correlation.} 
We further measured the correlation between WER/MR-WER vs. TER/WERd. We first calculated the scores for WER/MR-WER/TER/WERd for each of the 1,217 test utterances in isolation, and then we calculated the Pearson correlation using the corresponding lists of utterance-level scores. The results are shown in Table~\ref{tab:Pearson}. We can see that WERd correlates better than TER with both WER and also with MR-WER.
Overall, we can conclude that WERd is a promising measure for evaluating ASR systems that target dialectal speech input.

\begin{table}[h!]
\begin{center}
  \begin{tabular}{lc}
  \bf Metrics Compared & \bf Correlation \\
  \hline
  WER vs. TER  & 0.44 \\
  WER vs. WERd & 0.47 \\
  \hline
  MR-WER vs. TER  & 0.36 \\
  MR-WER vs. WERd & 0.39 \\
  \hline
  \end{tabular}
 \caption{Pearson correlations.}
 \label{tab:Pearson}
\end{center}
\end{table}







\smallskip
\noindent
\textbf{Closer look at the spelling variants used in test.}
Finally, we had a closer look at the 865 spelling variant pairs that were matched and used when calculating WERd for the test set of 1,217 segments, when using edit distance of 0.6. Our analysis shows three types of word-level changes:

\begin{enumerate}
  \item \emph{Word splitting}: 3\% of the pairs\\ e.g., \<مفيش> (mfy\$) $\rightarrow$ \<ما فيش> (mA fy\$).
  \item \emph{Word merging}: 16\% of the pairs\\ e.g., \<زي ما حنا> (zy mA HnA) $\rightarrow$ \<زي ماحنا> (zy mAHnA).
    \item \emph{Word substitution}: 81\% of the pairs\\ e.g., \<الامريكان> (AlAmrykAn) $\rightarrow$ \<الاميركان> (AlAmyrkAn).
\end{enumerate}

These statistics show that we learn many useful spelling variants, i.e., more than 80\%, rather than just splitting and merging words. 
Moreover, note that these word-level substitutions are actually small character-level transformations inside words.
Tables~\ref{tab:good_examples} and \ref{tab:bad_examples} show some examples of correct and wrong spelling variant pairs that were matched when calculating WERd for our Dialectal Arabic test set.

Table~\ref{tab:test_examples} further shows how spelling variant pairs affect hypothesis scoring for an example test sentence. There are three spelling variant pairs that match the input ASR hypothesis \colorbox{yellow}{\<ما فيش>} $\rightarrow$ \colorbox{yellow}{\<مفيش>},  \colorbox{green}{\<الامريكيه>} $\rightarrow$ \colorbox{green}{\<الاميركيه>}, and finally \colorbox{pink}{\<علشان>} $\rightarrow$ \colorbox{pink}{\<عشان>}. The former involves word splitting, while the latter two are about substitution. We can see that the WER after using spelling variant substitutions would go down to 30\%, (the actual WERd score would be slightly higher as it needs to take the cost of the spelling variant substitutions into account), while the initial WER was 61.5\%.

\begin{table}[tbh]
\small
\centering
\scalebox{0.85}{
\begin{tabular}{lccc}
\bf English Gloss & \multicolumn{2}{c}{\bf Spelling Variants} & \bf Operation \\
\hline
Netanyahu & \<نتانياهو> & ntAnyAhw & word substitution\\
& \<نتنياهو> & ntnyAhw &  \\
\hline
as we are & \<زي ما حنا> & zy mA HnA & word merging \\
& \<زي ماحنا> & zy mAHnA &  \\
\hline
talking & \<بتتكلم> & bttklm & word substitution \\
& \<تتكلم> & ttklm& \\
\hline
like (as if) & \<يعني> & yEny & word splitting\\
& \<يعني ب> & yEny b &  \\
\hline
\end{tabular}
}
\caption{\label{tab:good_examples} Correctly accepted spelling variants in test.}
\end{table}



\begin{table}[!htbp]
\small
\centering
\scalebox{0.85}{
\begin{tabular}{lccc}
\bf English Gloss & \multicolumn{2}{c}{\bf Spelling Variants} & \bf Operation \\
\hline
some & \<بعد> & bEd & word substitution \\
after & \<بعض> & bED &  \\
\hline
principal & \<رئيسي> & r\}ysy & word substitution\\
president &  \<رئيس> & r\}ys & \\
\hline
\end{tabular}
}
\caption{\label{tab:bad_examples} Wrongly accepted spelling variants in test.}
\end{table}

\begin{table}[!htbp]
\small
\centering
\scalebox{0.78}{
\begin{tabular}{l}
\hline
\textbf{Hypothesis (before)}: \<مفيش هم من مصر من الولايات المتحده الاميركيه عشان> \\
     \colorbox{yellow}{mfy\$} hm mn mSr mn AlwlAyAt AlmtHdh \colorbox{green}{AlAmyrkyh} \colorbox{pink}{E\$An} \\
WER: 61.54 [ 8/13; 0 insertions, 4 deletions, 4 substitutions ] \\
\hline
\textbf{Hypothesis (after)}: \<ما فيش هم من مصر من الولايات المتحده الامريكيه علشان> \\
      \colorbox{yellow}{mA fy\$} hm mn mSr mn AlwlAyAt AlmtHdh \colorbox{green}{AlAmrykyh} \colorbox{pink}{El\$An} \\
WER: 30.77 [ 4/13; 0 insertions, 3 deletions, 1 substitutions ] \\
\hline
\textbf{Reference}: \<ما فيش زيهم جم من مصر وجم من كل الولايات المتحده الامريكيه علشان> \\
     \colorbox{yellow}{mA fy\$} zyhm jm mn mSr wjm mn kl AlwlAyAt AlmtHdh \colorbox{green}{AlAmrykyh} \colorbox{pink}{El\$An} \\      
\hline
\end{tabular}
}
\caption{\label{tab:test_examples}
Extra word matches due to using spelling variants. Shown is an ASR hypothesis for a test utterance, and the impact of hypothesis matching on the number of insertions, deletions and substitutions, as well as on the overall WER score.}
\end{table}

\section{Conclusion and Future Work}

We have addressed the evaluation of ASR systems that target dialectal speech input, where a major problem is the natural variation in spelling due to the unofficial status and the lack of standardization of the orthography.
We have proposed a new metric, WERd (or \emph{WER for dialects}), a variation of TERp, for which multiple text outputs for the same speech signal can be acceptable given a single reference transcript. Our implementation of WERd was based on mining 11M pairs of spelling variants from a huge dialectal Arabic tweet collection. Our automatic experiments, as well as manual analysis, have shown that this is a highly promising metric that addresses the problems of WER for dialectal speech, and approaches the performance of  multi-reference WER.

In future work, we plan experiments with other dialects and non-standardized language varieties. We also want to incorporate word embeddings in the process of computation, e.g., character-based, which can naturally tolerate some spelling variation \cite{BojanowskiGJM17}. We further want to explore using weighted finite state transducers as an alternative way to allow using multiple spelling variants for both references and hypotheses. 

Last but not least, we have made publicly available our code together with our data and the spelling variants we have mined. We hope that this will enable further research in ASR evaluation for languages with non-standardized orthography. 

\section{Acknowledgements}


This work was partially supported by the EU H2020 project SUMMA, under grant agreement 688139.

\bibliographystyle{IEEEbib}
\bibliography{refs}

\begin{thebibliography}{10}

\bibitem{saon2017english}
George Saon, Gakuto Kurata, Tom Sercu, Kartik Audhkhasi, Samuel Thomas,
  Dimitrios Dimitriadis, Xiaodong Cui, Bhuvana Ramabhadran, Michael Picheny,
  Lynn-Li Lim, Bergul Roomi, and Phil Hall,
\newblock ``English conversational telephone speech recognition by humans and
  machines,''
\newblock {\em arXiv preprint arXiv:1703.02136}, 2017.

\bibitem{aliMGB3}
Ahmed Ali, Stephan Vogel, and Steve Renals,
\newblock ``Speech {R}ecognition {C}hallenge in the {W}ild: {A}rabic {MGB}-3,''
\newblock in {\em Proceedings of the IEEE Workshop on Automatic Speech
  Recognition and Understanding}, Okinawa, Japan, 2017, ASRU~'17.

\bibitem{rosenfeld1997cmu}
Ronald Rosenfeld and Philip Clarkson,
\newblock ``{CMU-Cambridge} statistical language modeling toolkit v2,'' 1997.

\bibitem{fiscus1997post}
Jonathan Fiscus,
\newblock ``A post-processing system to yield reduced word error rates:
  Recognizer output voting error reduction ({ROVER}),''
\newblock in {\em Proceedings of the IEEE Workshop on Automatic Speech
  Recognition and Understanding}, Santa Barbara, California, USA, 1997,
  ASRU~'97, pp. 347--354.

\bibitem{ali2015multi}
Ahmed Ali, Walid Magdy, Peter Bell, and Steve Renals,
\newblock ``Multi-reference {WER} for evaluating {ASR} for languages with no
  orthographic rules,''
\newblock in {\em Proceedings of the IEEE Workshop on Automatic Speech
  Recognition and Understanding}, Scottsdale, Arizona, USA, 2015, ASRU~'15, pp.
  576--580.

\bibitem{Papineni:2002:BLEU}
Kishore Papineni, Salim Roukos, Todd Ward, and Wei-Jing Zhu,
\newblock ``{BLEU}: A method for automatic evaluation of machine translation,''
\newblock in {\em Proceedings of the 40th Annual Meeting on Association for
  Computational Linguistics}, Philadelphia, Pennsylvania, USA, 2002, ACL~'02,
  pp. 311--318.

\bibitem{madnani-EtAl:2007:WMT}
Nitin Madnani, Necip Fazil~Ayan, Philip Resnik, and Bonnie Dorr,
\newblock ``Using paraphrases for parameter tuning in statistical machine
  translation,''
\newblock in {\em Proceedings of the Second Workshop on Statistical Machine
  Translation}, Prague, Czech Republic, 2007, WMT~'07, pp. 120--127.

\bibitem{Callison-Burch:al:2006:mt}
Chris Callison-Burch, Philipp Koehn, and Miles Osborne,
\newblock ``Improved statistical machine translation using paraphrases,''
\newblock in {\em Proceedings of the Conference on Human Language Technology
  Conference of the North American Chapter of the Association for Computational
  Linguistics}, New York, New York, USA, 2006, NAACL-HLT~'06, pp. 17--24.

\bibitem{nakov:2008:WMT}
Preslav Nakov,
\newblock ``Improving {English-Spanish} statistical machine translation:
  Experiments in domain adaptation, sentence paraphrasing, tokenization, and
  recasing,''
\newblock in {\em Proceedings of the Third Workshop on Statistical Machine
  Translation}, Columbus, Ohio, USA, 2008, WMT~'08, pp. 147--150.

\bibitem{Nakov:2008:ISM}
Preslav Nakov,
\newblock ``Improved statistical machine translation using monolingual
  paraphrases,''
\newblock in {\em Proceedings of the 18th European Conference on Artificial
  Intelligence}, Patras, Greece, 2008, ECAI~'08, pp. 338--342.

\bibitem{Nakov:2011:TMC}
Preslav Nakov and Hwee~Tou Ng,
\newblock ``Translating from morphologically complex languages: A
  paraphrase-based approach,''
\newblock in {\em Proceedings of the 49th Annual Meeting of the Association for
  Computational Linguistics: Human Language Technologies}, Portland, Oregon,
  USA, 2011, ACL~'11, pp. 1298--1307.

\bibitem{Wang:2012:SLA}
Pidong Wang, Preslav Nakov, and Hwee~Tou Ng,
\newblock ``Source language adaptation for resource-poor machine translation,''
\newblock in {\em Proceedings of the 2012 Joint Conference on Empirical Methods
  in Natural Language Processing and Computational Natural Language Learning},
  Jeju Island, Korea, 2012, EMNLP-CoNLL~'12, pp. 286--296.

\bibitem{Wang:2016:SLA}
Pidong Wang, Preslav Nakov, and Hwee~Tou Ng,
\newblock ``Source language adaptation approaches for resource-poor machine
  translation,''
\newblock {\em Comput. Linguist.}, vol. 42, no. 2, pp. 277--306, June 2016.

\bibitem{Zhou:al:2006:mt}
Liang Zhou, Chin-Yew Lin, and Eduard Hovy,
\newblock ``Re-evaluating machine translation results with paraphrase
  support,''
\newblock in {\em Proceedings of the Conference on Empirical Methods in Natural
  Language Processing}, Sydney, Australia, 2006, EMNLP~'06, pp. 77--84.

\bibitem{Kauchak:Barzilay:2006:par}
David Kauchak and Regina Barzilay,
\newblock ``Paraphrasing for automatic evaluation,''
\newblock in {\em Proceedings of the Conference on Human Language Technology
  Conference of the North American Chapter of the Association for Computational
  Linguistics}, New York, New York, USA, 2006, NAACL-HLT~'06, pp. 455--462.

\bibitem{Snover:2009:TERp}
Matthew~G. Snover, Nitin Madnani, Bonnie Dorr, and Richard Schwartz,
\newblock ``{TER-Plus}: Paraphrase, semantic, and alignment enhancements to
  translation edit rate,''
\newblock {\em Machine Translation}, pp. 117--127, 2009.

\bibitem{darwish2012language}
Kareem Darwish, Walid Magdy, and Ahmed Mourad,
\newblock ``Language processing for {A}rabic microblog retrieval,''
\newblock in {\em Proceedings of the 21st ACM International Conference on
  Information and Knowledge Management}, Maui, Hawaii, USA, 2012, CIKM~'12, pp.
  2427--2430.

\bibitem{hassan2013social}
Hany Hassan and Arul Menezes,
\newblock ``Social text normalization using contextual graph random walks,''
\newblock in {\em Proceedings of the Annual Meeting on Association for
  Computational Linguistics}, Sofia, Bulgaria, 2013, ACL~'13, pp. 1577--1586.

\bibitem{han2012automatically}
Bo~Han, Paul Cook, and Timothy Baldwin,
\newblock ``Automatically constructing a normalisation dictionary for
  microblogs,''
\newblock in {\em Proceedings of the Joint Conference on Empirical Methods in
  Natural Language Processing and Computational Natural Language Learning},
  Jeju Island, Korea, 2012, EMNLP-CoNLL~'12, pp. 421--432.

\bibitem{sridhar2015unsupervised}
Vivek~Kumar Rangarajan~Sridhar,
\newblock ``Unsupervised text normalization using distributed representations
  of words and phrases,''
\newblock in {\em Proceedings of the 1st Workshop on Vector Space Modeling for
  Natural Language Processing}, Denver, Colorado, USA, 2015, pp. 8--16.

\bibitem{sproat2016rnn}
Richard Sproat and Navdeep Jaitly,
\newblock ``{RNN} approaches to text normalization: A challenge,''
\newblock {\em arXiv preprint arXiv:1611.00068}, 2016.

\bibitem{snover2006study}
Matthew Snover, Bonnie Dorr, Richard Schwartz, Linnea Micciulla, and John
  Makhoul,
\newblock ``A study of translation edit rate with targeted human annotation,''
\newblock in {\em Proceedings of Association for Machine Translation in the
  Americas}, Cambridge, Massachusetts, USA, 2006, AMTA~'06.

\bibitem{wray2015crowdsource}
Samantha Wray and Ahmed Ali,
\newblock ``Crowdsource a little to label a lot: Labeling a speech corpus of
  dialectal {A}rabic,''
\newblock in {\em Proceedings of the 16th Annual Conference of the
  International Speech Communication Association}, Dresden, Germany, 2015,
  INTERSPEECH~'15, pp. 2824--2828.

\bibitem{khurana2016qcri}
Sameer Khurana and Ahmed Ali,
\newblock ``{QCRI} advanced transcription system ({QATS}) for the {A}rabic
  multi-dialect broadcast media recognition: {MGB}-2 challenge,''
\newblock in {\em Proceedings of the IEEE Workshop on Spoken Language
  Technology}, San Diego, California, USA, 2016, SLT~'16, pp. 292--298.

\bibitem{ali2016mgb}
Ahmed Ali, Peter Bell, James Glass, Yacine Messaoui, Hamdy Mubarak, Steve
  Renals, and Yifan Zhang,
\newblock ``The {MGB}-2 challenge: {A}rabic multi-dialect broadcast media
  recognition,''
\newblock in {\em Proceedings of the IEEE Workshop on Spoken Language
  Technology}, San Diego, California, USA, 2016, SLT~'2016, pp. 279--284.

\bibitem{ko2015audio}
Tom Ko, Vijayaditya Peddinti, Daniel Povey, and Sanjeev Khudanpur,
\newblock ``Audio augmentation for speech recognition,''
\newblock in {\em Proceedings of the 16th Annual Conference of the
  International Speech Communication Association}, Dresden, Germany, 2015,
  INTERSPEECH~'15, pp. 3586--3589.

\bibitem{killer2003grapheme}
Mirjam Killer, Sebastian St{\"u}ker, and Tanja Schultz,
\newblock ``Grapheme based speech recognition,''
\newblock in {\em Proceedings of the 8th European Conference on Speech
  Communication and Technology}, Geneva, Switzerland, 2003,
  EUROSPEECH--INTERSPEECH~'03.

\bibitem{saon2013speaker}
George Saon, Hagen Soltau, David Nahamoo, and Michael Picheny,
\newblock ``Speaker adaptation of neural network acoustic models using
  i-vectors,''
\newblock in {\em Proceedings of the IEEE Workshop on Automatic Speech
  Recognition and Understanding}, Olomouc, Czech Republic, 2013, ASRU~'13, pp.
  55--59.

\bibitem{peddinti2015time}
Vijayaditya Peddinti, Daniel Povey, and Sanjeev Khudanpur,
\newblock ``A time delay neural network architecture for efficient modeling of
  long temporal contexts,''
\newblock in {\em Proceedings of the 16th Annual Conference of the
  International Speech Communication Association}, Dresden, Germany, 2015,
  INTERSPEECH~'15, pp. 3214--3218.

\bibitem{sak2014long}
Ha{\c{s}}im Sak, Andrew Senior, and Fran{\c{c}}oise Beaufays,
\newblock ``Long short-term memory recurrent neural network architectures for
  large scale acoustic modeling,''
\newblock in {\em Proceedings of the 15th Annual Conference of the
  International Speech Communication Association}, Singapore, 2014,
  INTERSPEECH~'14, pp. 338--342.

\bibitem{povey2016purely}
Daniel Povey, Vijayaditya Peddinti, Daniel Galvez, Pegah Ghahremani, Vimal
  Manohar, Xingyu Na, Yiming Wang, and Sanjeev Khudanpur,
\newblock ``Purely sequence-trained neural networks for {ASR} based on
  lattice-free {MMI},''
\newblock in {\em Proceedings of the 17th Annual Conference of the
  International Speech Communication Association}, San Francisco, California,
  USA, 2016, INTERSPEECH~'16, pp. 2751--2755.

\bibitem{povey2011kaldi}
Daniel Povey, Arnab Ghoshal, Gilles Boulianne, Lukas Burget, Ondrej Glembek,
  Nagendra Goel, Mirko Hannemann, Petr Motlicek, Yanmin Qian, Petr Schwarz,
  et~al.,
\newblock ``The {K}aldi speech recognition toolkit,''
\newblock in {\em Proceedings of the IEEE Workshop on Automatic Speech
  Recognition and Understanding}, Waikoloa, Hawaii, USA, 2011, ASRU~'11.

\bibitem{stolcke2002srilm}
Andreas Stolcke et~al.,
\newblock ``{SRILM} - an extensible language modeling toolkit,''
\newblock in {\em Proceedings of the 7th International Conference on Spoken
  Language Processing}, Denver, Colorado, USA, 2002, ICSLP-INTERSPEECH~'02.

\bibitem{mikolov2011rnnlm}
Tomas Mikolov, Stefan Kombrink, Anoop Deoras, Lukar Burget, and Jan Cernocky,
\newblock ``{RNNLM} - recurrent neural network language modeling toolkit,''
\newblock in {\em Proceedings of the IEEE Workshop on Automatic Speech
  Recognition and Understanding}, Big Island, Hawaii, USA, 2011, ASRU~'11, pp.
  196--201.

\bibitem{BojanowskiGJM17}
Piotr Bojanowski, Edouard Grave, Armand Joulin, and Tomas Mikolov,
\newblock ``Enriching word vectors with subword information,''
\newblock {\em {TACL}}, vol. 5, pp. 135--146, 2017.

\end{thebibliography}

\end{document}